\documentclass[letterpaper, 10 pt, conference]{ieeeconf}  

\IEEEoverridecommandlockouts                              
\usepackage{amsmath}
\usepackage{graphicx}
\usepackage[caption=false]{subfig}
\usepackage{multirow}
\usepackage{dblfloatfix}
\usepackage{siunitx}
\usepackage{booktabs}
\usepackage{longtable}

\title{\LARGE \bf
CobotGear: Interaction with Collaborative Robots using
 \\ Wearable Optical Motion Capturing Systems  
}

\author{Juan Heredia,
Miguel Altamirano Cabrera, 
Jonathan Tirado, \\ 
Vladislav Panov,
 Dzmitry Tsetserukou
\thanks{The authors are with the Space Center, Skolkovo Institute of Science and Technology (Skoltech), 121205 Bolshoy Boulevard 30, bld. 1, Moscow, Russia. {\tt\small \{juan.heredia, miguel.altamirano, jonathan.tirado, vladislav.panov, d.tsetserukou\}@skoltech.ru\ }}}

\begin{document}

\maketitle
\thispagestyle{empty}
\pagestyle{empty}

\begin{abstract}

In industrial applications, complex tasks require human collaboration since the robot doesn't have enough dexterity. However, the robots are still implemented as tools and not as collaborative intelligent systems. To ensure safety in the human-robot collaboration, we introduce a system that presents a new method that integrates low-cost wearable mocap, and an improved collision avoidance algorithm based on the artificial potential fields. Wearable optical motion capturing allows to track the human hand position with high accuracy and low latency on large working areas. To increase the efficiency of the proposed algorithm, two obstacle types are discriminated according to their collision probability. A preliminary experiment was performed to analyze the algorithm behavior and to select the best values for the obstacle's threshold angle $\theta_{OBS}$, and for the avoidance threshold distance $d_{AT}$. The second experiment was carried out to evaluate the system performance with $d_{AT}$ = 0.2 m and $\theta_{OBS}$ = 45 degrees. The third experiment evaluated the system in a real collaborative task. The results demonstrate the robust performance of the robotic arm generating smooth collision-free trajectories. The proposed technology will allow consumer robots to safely collaborate with humans in cluttered environments, e.g., factories, kitchens, living rooms, and restaurants.

\end{abstract}

\section{Introduction}

The implementation of robots in industrial and service sectors has been increasing in the last years. However, in the majority of domains, robots are still implemented more as tools than as intelligent systems that can collaborate with humans. Close collaboration between robots and humans requires a flexible, fast, and safety-oriented control capable of ensuring human protection in dynamically changing environments \cite{Bicchi2008}. 

\begin{figure}[h!]
  \centering
  \includegraphics[width=0.43\textwidth]{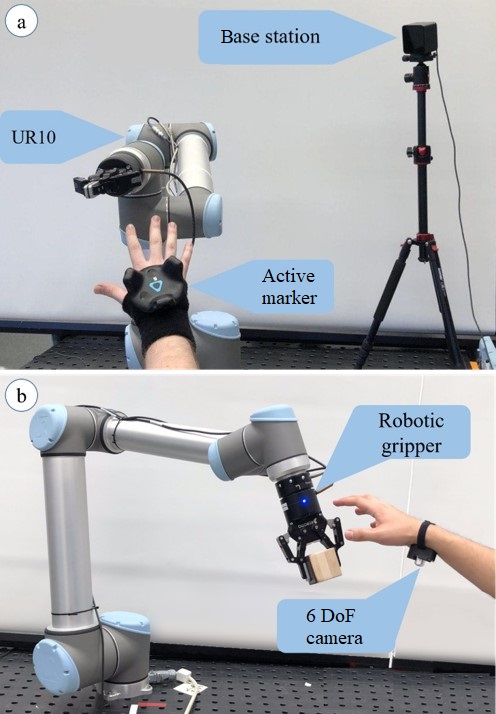}
  \qquad
  \caption{Layout of the collaborative robotic system CobotGear. a) Collision avoidance using the HTC technology. b) Collision avoidance using the AntiLatency technology. }
  \label{fig:SystemOverview}
\end{figure}

To provide the human-robot collaboration, the robot is equipped with the sensors to detect the collision with humans. For example, force-torque sensors have been used at the end-effector \cite{Siciliano2019}, or at the base \cite{Das}. In \cite{Tsetserukou}, an obstacle avoidance control through tactile perception using optical torque sensors was introduced. However, the interaction with the environment is limited due to the contact of the robotic arm with obstacles.  Because all off-the-shelf collaborative robots are equipped only with force/torque sensors, they stop after the collision is detected. High-payload robots present danger to human during physical contact. Therefore, to guarantee a high level of safety to human we must generate the collision-free trajectories in advance to avoid any contact with the human. In \cite{MColo2020}, the use of a capacitive skin was suggested, which was located on the links of a 6 DOF collaborative robot for measuring the distance between the robot's Tool Center Point (TCP) and the obstacles to avoid. However, the capacitive skin sensors are costly and were installed to detect objects at close distances. Ding et al. \cite{Ding2020} proposed the use of multi-modal proximity sensor modules on the robot links and TCP to calculate the obstacle position. Nevertheless, the sensors have to be added to the tool to perceive the obstacle location. 

The environment perception is used to detect the real-time position of the obstacles to avoid. The implementation of cameras and computer vision algorithms for obstacle avoidance has been done in \cite{Cuevas-Velasquez2018}, providing a system for hybrid visual servoing to moving targets. However, the system's complexity requires careful calibration, and potentially, the data acquisition can be affected by latency due to the long processing time of each image frame. 
There are many environment perception systems based on different techniques in the market. Nowadays, passive optical mocap, such as VICON, OptiTrack, or NDI's Polaris, is the most popular solution. Their advantages are low latency and high precision. However, such systems require a powerful computer for data processing, have a limited working area, and are very expensive. Additionally, passive markers can be occluded by the robot moving above the human hand. We propose to apply 6 DOF wearable optical mocap (by AntiLatency) featuring onboard data processing with FPGA resulting in low latency of 2 ms. Importantly, the required working area can be easily arranged by placing cheap IR markers on the floor. In the case wearable camera is temporally occluded, the hand position is available thanks to the IMU odometry.   

The Artificial Potential Field (APF) approach, introduced in \cite{Khatib}, proposes virtual repulsive and attractive fields associated with obstacles and targets faced during the robot movement. This approach was implemented in autonomous mobile robots. It was further elaborated in \cite{SGe}, introducing new repulsive potential functions to avoid the local minimum problem, ensuring that the robot can reach the goal. In \cite{HLin}, a repulsive force field in three-dimensional space was introduced to adjust the original trajectory of the robot and avoid obstacles smoothly. The idea of using potential field with a fictional repulsive force around the workspace obstacles was suggested in \cite{Liao} and later developed in \cite{Luecke}.

This paper presents a novel collision avoidance method for collaborative robots composed of the essential elements introduced in \cite{Flacco2012}. A behavior tree chooses the control technique between collision avoidance algorithms and robot position control algorithm. We propose a novel collision avoidance algorithm based on artificial potential fields. For experimental evaluation, the designed work cell prototype has the following components:
\begin{itemize}
  \item Collaborative robot UR10.
  \item Optical active motion capture device HTC VR VIVE.
  \item Wearable mocap by AntiLatency. 
\end{itemize}

\section{Proposed Approach}

\subsection{Environment Perception}

As was mentioned in Section I, active optical technology has a low latency, and its systems are cheap and easy to scale. 
For that reason, we selected HTC VR VIVE and AntiLatency Technology. 

HTC VIVE technology uses two base stations (light-houses) as references. The tracker has 18 infrared sensors that receive the signal from base stations. The position of the device is determined by the time difference between the signals emitted by each base. This system has a low latency. However, it requires base stations and has glitches caused by reflective surfaces in the room. Therefore, HTC VIVE is not suitable for industrial environments. We used this tracking technology for preliminary experiments of the algorithm. 

AntiLatency technology is a novel optical mocap system. Reference stripes with active infrared markers are located on the floor and a 6 DOF wearable optical camera is attached to the tracked object. The device calculated the position and orientation with onboard FPGA. For detecting the markers on the floor, the AntiLatency tracker uses a camera. While being temporally occluded, the hand position is calculated by IMU odometry. We propose to apply this novel technology, which has been used only in virtual reality environments, to the industry. This technology was used in the final experiment to demonstrate the feasibility of using wearable mocap in an industrial environment with a real task.

\subsection{Control Algorithm}
The designed control algorithm aims to converge the robot TCP to the desired position. At the same time, the operator position is tracked, and the robot avoids the collision. In this article, a collaborative robot of 6 DOF is considered. Let $\boldsymbol{x_{R}}$,  and ${q}$ be the Cartesian position of the robot TCP and the position of each joint, respectively. The relation between the Cartesian velocity and each joint speed is described by the analytic Jacobian $J$ using the equation: $\dot{q} = J^{-1} \dot{v}$.

 \begin{figure}[t!]
  \centering
  \includegraphics[width=0.47\textwidth]{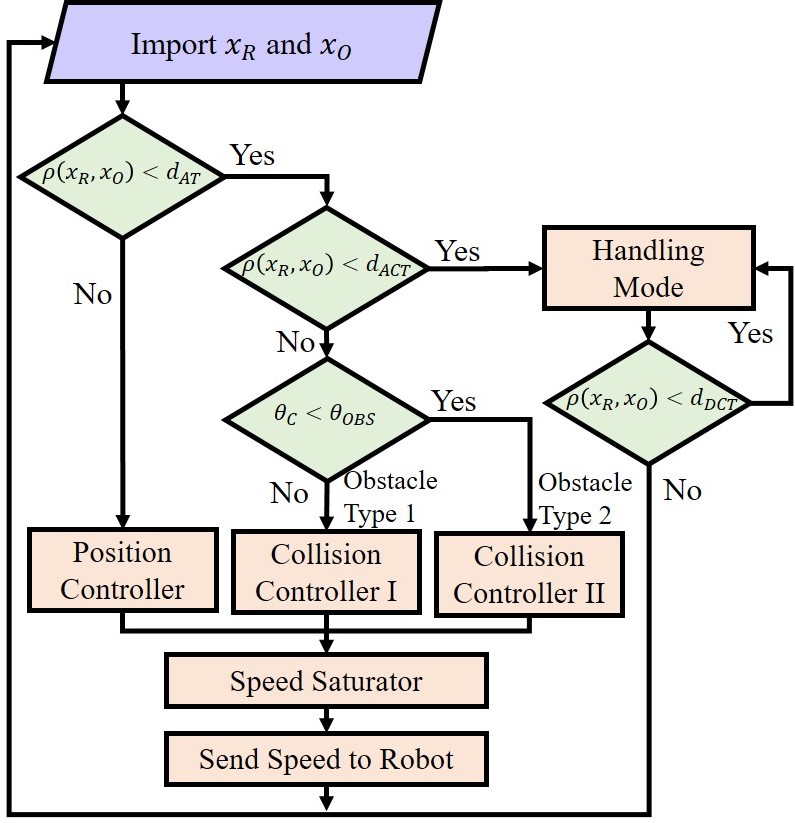}
  \qquad
  \caption{Behavior tree of the control architecture. Position controller and collision controller I and II are defined by (\ref{eq:C}), (\ref{eq:CI}), and (\ref{eq:CII}), respectively. } 
  \label{fig:BT}
\end{figure}

To address our problem, we propose a behavior tree, represented as a diagram in Fig. \ref{fig:BT}. The decisions are taken based on the distance between the robot TCP and the obstacle $d_{RO} = \rho(\boldsymbol{x}_{R},\boldsymbol{x}_{O}) $. When $d_{RO}$  is bigger than the avoidance threshold distance $d_{AT}$, the algorithm deploys the position controller that is represented in (\ref{eq:C}). When  $d_{RO}$ is smaller than the critical threshold distance of activation $d_{ACT}$, the robot initiates a free drive control mode. In the free drive control mode, the user can modify the robot's position by pulling-pushing end-effector. To increase the system safety, the condition $d_{DCT} > d_{ACT}$ is proposed. The system deactivates the free drive control mode when the robot-operator distance $d_{RO}$ is bigger than the critical threshold of deactivation $d_{DCT}$. The parameter $d_{DCT}$ is configured to guarantee operator safety. When the TCP is inside of the avoidance area ($d_{RO} < d_{AT}$ and $d_{RO} > d_{ACT}$), the algorithm classifies the obstacle as two different types. Equations (\ref{eq:CI}) and (\ref{eq:CII}) refer to the avoidance control in the case of obstacle type 1 and obstacle type 2, respectively. 
 
\subsubsection{Position Controller}

the position controller aiming to achieve the goal position ${x_{G}}$ without collision avoidance is defined as follow: 
\begin{equation} \boldsymbol{e} = \boldsymbol{x}_{R} - \boldsymbol{x}_{G} \end{equation}
\begin{equation} \boldsymbol{v}_{PC}= ( \boldsymbol{\dot{x}}_{G} + k_{PC1} \tanh{(k_{PC2} \cdot \boldsymbol{e})}) \end{equation}
\begin{equation}\label{eq:C} \boldsymbol{\dot{q}}_{PC} = J^{-1} \cdot ( \boldsymbol{\dot{x}}_{G} + k_{PC1} \tanh{(k_{PC2} \cdot \boldsymbol{e})}), \end{equation}
where $\boldsymbol{e}$ is the position error, the parameters $\boldsymbol{v}_{PC}$ and $\boldsymbol{\dot{q}}_{PC}$ are the Cartesian velocity and the angular velocity of each motor, respectively. The parameters $k_{PC1}$ and $k_{PC2}$ are the calibration constants of the Position Controller Algorithm. 

\subsubsection{Collision Avoidance}
\begin{figure}[h!]
  \centering
  \includegraphics[width=0.45\textwidth]{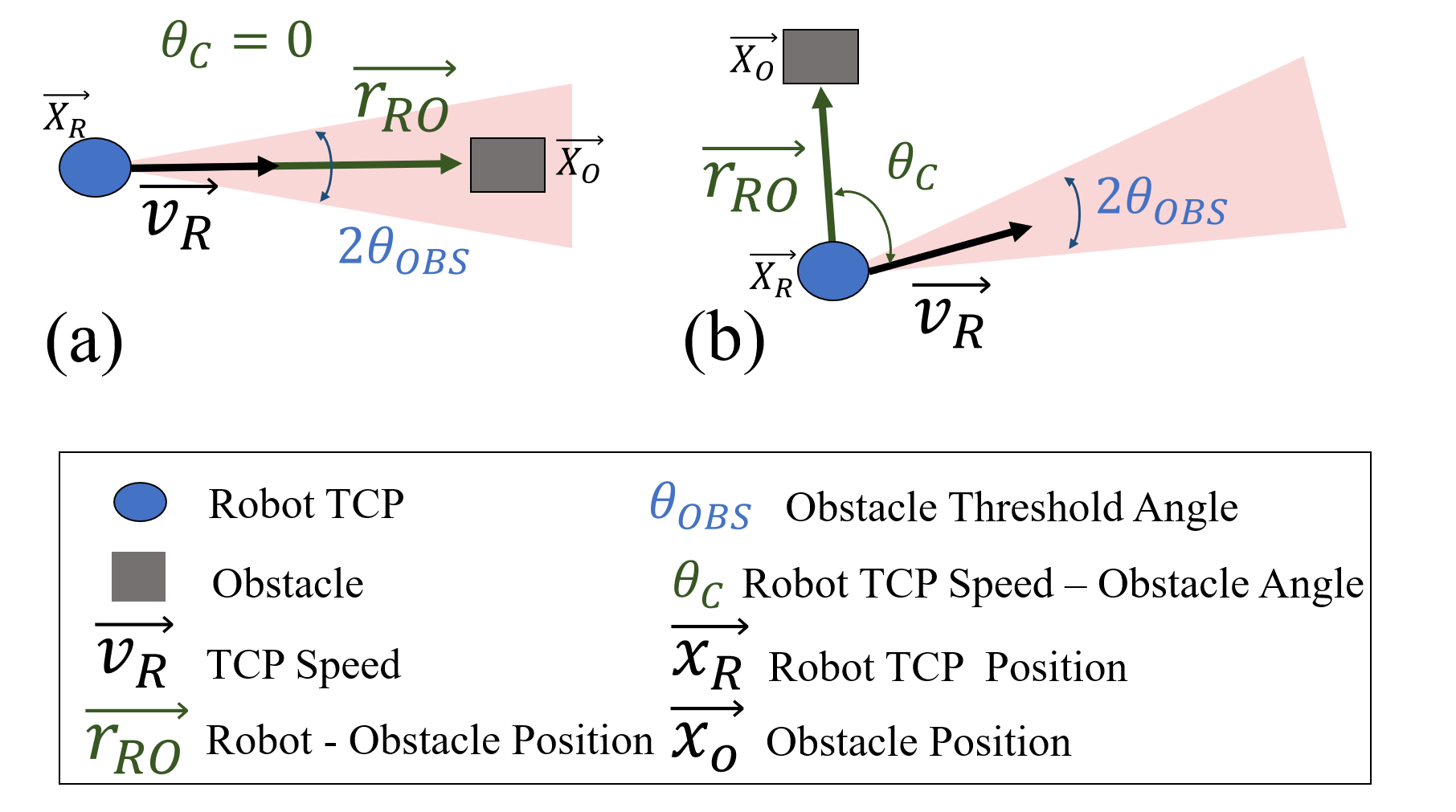}
  \qquad
  \caption{a) Obstacle Type 1: imminent collision. The angle between the Robot-Obstacle vector and robot velocity is smaller than $\theta_{OBS}$. b) Obstacle Type 2: no imminent collision. The angle between the Robot-Obstacle vector and robot velocity is bigger than $\theta_{OBS}$. }
  
  \label{fig:vectors}
\end{figure}

we propose an algorithm based on APF. The algorithm discriminates two types of obstacles, as shown in Fig. \ref{fig:vectors}, aiming to increase the efficiency of traditional approaches. As shown in \cite{7334399}, the probability of collision is inversely proportional to the distance between the robot and the human hand.

\textbf{Obstacle Type 1:} it is an imminent barrier, and the probability that the robot collides with the obstacle is high. To consider an obstacle as type 1, the angle between the TCP velocity and the Robot-Obstacle vector should be smaller than $\theta_{OBS}$. The constant $\theta_{OBS}$ is a parameter of calibration. To avoid this type of obstacle, the algorithm is composed of a common repulsive force and two rotational repulsive forces. 

The first repulsive force follows the traditional approach:
\begin{equation}
\boldsymbol{F}_{rep1} = -  \dfrac{k_{CA1}}{\rho(\boldsymbol{x}_{R}, \boldsymbol{x}_{O})^2} \boldsymbol{n}_{RO}, 
\end{equation}
where function $\rho(\boldsymbol{x}_{1}, \boldsymbol{x}_{2})$ represents the Cartesian distance between two vectors $\boldsymbol{x}_{1}$ and $\boldsymbol{x}_{2}$, $k_{CA1}$ represents a calibration constant for the first repulsive force, and  $\boldsymbol{n}_{RO}$ is the unit vector of the Robot-Obstacle vector.

The second and third repulsive forces are rotational. Applying these forces, the robot avoids the obstacle with a soft circular trajectory. The forces must be perpendicular to the Robot-Obstacle vector $\boldsymbol{x}_{RO}$. There is an infinite number of vectors that satisfy the previous description; in this paper, we propose to use two of them. Specifically, vectors $\boldsymbol{a}$ and $\boldsymbol{b}$ are auxiliary vectors used to generate the direction of the two rotational repulsive forces: 

\begin{equation} \boldsymbol{a}_1 = \boldsymbol{v}_{R} \times \boldsymbol{n}_{RO} 
\end{equation}
\begin{equation}
\boldsymbol{b}_1 = \boldsymbol{a}_1 \times \boldsymbol{n}_{RO} 
\end{equation}
\begin{equation} \boldsymbol{F}_{rep2} = \dfrac{k_{CA2} \cdot c_1 }{\rho(\boldsymbol{x}_{R},\boldsymbol{x}_{O})^2} \boldsymbol{n}_{b_1},\end{equation}
where $k_{CA2}$ is the collision avoidance calibration constant for the second repulsive force, $\boldsymbol{n}_{b_1}$ is the unit vector of the vector $\boldsymbol{b_1}$, and $c_1$ is the direction vector constant.%

Compared to the working area of a mobile robot, the working area of a collaborative robot is significantly smaller. When a mobile robot collision avoidance algorithm is applied to a manipulator, the robot has a high risk of entering into a singularity, pursuing positions outside of its working area. The proposed algorithm attempts to avoid singularities and crashes with environmental elements. Therefore, the rotational repulsive force direction must be oriented to the robot base. When the angle between the rotational force vector ($\boldsymbol{n}_{b_1}$) and the robot base - TCP position vector ($-x_R$) is smaller than $\frac{\pi}{2}$, $c_1 $ is equal to $1$, otherwise $c_1 $ is equal to $ -1$. 
\begin{equation}
\boldsymbol{a}_2 = \boldsymbol{n}_{RG} \times \boldsymbol{n}_{RO} 
\end{equation}
\begin{equation}
\boldsymbol{b}_2 = \boldsymbol{a}_2 \times \boldsymbol{n}_{RO} 
\end{equation}
\begin{equation}
\boldsymbol{F}_{rep3} = \dfrac{ k_{CA3} \cdot c_2  }{\rho(x_{R}, x_{O})^2} \boldsymbol{n}_{b_2}, 
\end{equation}
where $\boldsymbol{n}_{RG}$ is the unit vector of the TCP-Goal vector, $k_{CA3}$ is the collision avoidance calibration constant for the third repulsive force, $\boldsymbol{n}_{b_2}$ is the unit vector of the vector $\boldsymbol{b_2}$, and $c_2$ is the direction vector constant. When the angle between the rotational force vector $\boldsymbol{n}_{b_2}$ and the Robot Base - TCP position vector ($-x_R$) is smaller than $\frac{\pi}{2}$, $c_2 $ equals $ 1$, otherwise $c_2 $ equals $ -1$. \begin{equation}
\boldsymbol{F}_{rep} = \boldsymbol{F}_{rep1} +\boldsymbol{F}_{rep2} +\boldsymbol{F}_{rep3}
\end{equation}
\begin{equation}
\boldsymbol{v}_{rep} = \boldsymbol{v}_{R} + k_{rep} \cdot \boldsymbol{F}_{rep},  
\end{equation}
where $k_{rep}$ is the relation between the force and the variation of velocity. It means that $k_{rep} = T/m $, where $T$ is the sampling time, and $m$ is an artificial mass. However, the constant $k_{rep}$ can be included inside the parameter $k_{CAi}$ of each repulsive force.

The controller is composed of two components, the first one attempts to achieve the desired position, and the second avoids the environment obstacles. To combine two tasks, we use a null-space approach:
\begin{equation}
\label{eq:CI}
\dot{q} = J^{-1}( \boldsymbol{v}_{PC} ( 1 - e^{\tau \rho(x_{R}, r_{O}) }) + \boldsymbol{v}_{rep}  e^{\tau \rho(x_{R}, x_{O} )}), 
\end{equation}
where $\tau$ represents the space-null attenuation constant. 

 \textbf{Obstacle Type 2:} it is a no imminent barrier, and the probability that the robot collides with the obstacle is low. To consider an obstacle as type 2, the angle between the TCP velocity and the Robot-Obstacle vector is higher than $\theta_{OBS}$. The collision avoidance algorithm is based on the traditional approach of APF: 
\begin{equation} \boldsymbol{F}_{rep} = \boldsymbol{F}_{rep1} \end{equation}
\begin{equation} \boldsymbol{v}_{rep} = \boldsymbol{v}_{R} + k_{rep} \cdot \boldsymbol{F}_{rep}  \end{equation}
\begin{equation} \label{eq:CII} \dot{q} = J^{-1}( \boldsymbol{v}_{PC} ( 1 - e^{\tau \rho(x_{R}, r_{O}) }) + \boldsymbol{v}_{rep}  e^{\tau \rho(x_{R}, x_{O} )}). \end{equation}

The proposed algorithm needs to be evaluated in experiments of human-robot collaboration. 
\section{Algorithm Calibration Experiments} \label{ALGORITHM CALIBRATION EXPERIMENTS}

The following two experiments were designed to analyze the proposed algorithm's behavior over different conditions. Each experiment considered only one algorithm factor ($\theta_{OBS}$, or $d_{AT}$) as an independent variable.

\begin{figure}[h!]
  \centering
  \includegraphics[width=0.4\textwidth]{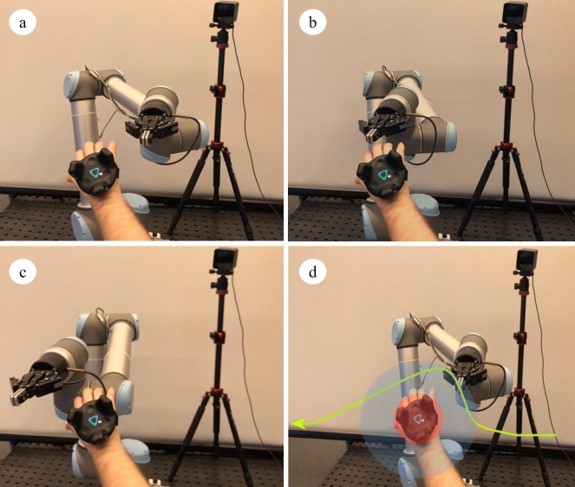}
  \qquad
  \caption{ Experiment for algorithm calibration. a) Initial position. b) Obstacle type 1. c) Obstacle type 2. d) Obstacle avoidance trajectory (yellow line). }
  \label{fig:Interaction}
\end{figure}

The experimental parameters were set up as follows.
As shown in Fig. \ref{fig:Interaction}, the robot moved between two points. The Initial point was located at the position (-0.3; 0.8; 0.7) m, and the Final at (0.3; 0.8; 0.7) m, both of them referenced to the robot's base. The maximum speed of the robot was set to 20 cm/sec. Critical threshold distance of activation $d_{ACT}$ and deactivation $d_{DCT}$ was fixed at 5 cm and 20 cm, respectively.  Space-null attenuation constant $\tau$ was set to 20 cm\textsuperscript{-1}. The sampling time for all of the experiments was 100 ms. The experiments were developed using the Universal Robot UR10 in a semi-structured environment.

\subsection{Obstacle's threshold angle $\theta_{OBS}$}

\begin{figure}[h]
  \centering
  \includegraphics[width=0.47\textwidth]{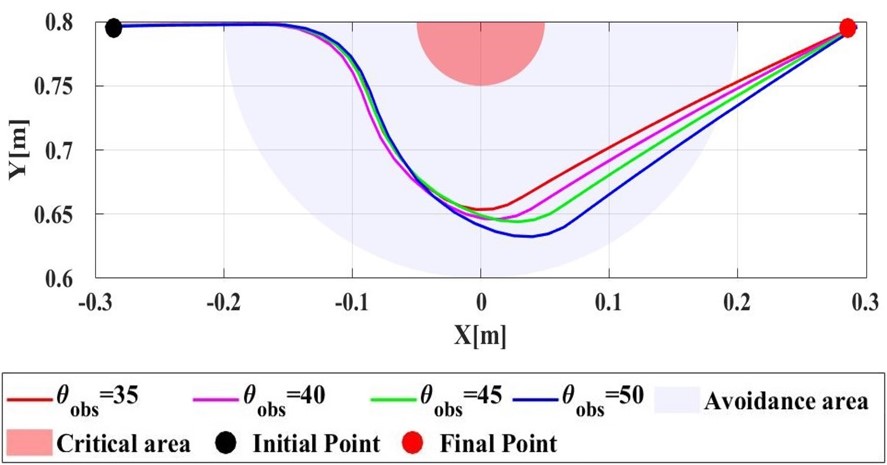}
  \qquad
  \caption{Experiment 1: the end-effector path curvature in relation to $\theta_{OBS}$ values, plane (XY) view.}
  \label{fig:kangle}
\end{figure}

For the first experiment, $\theta_{OBS}$ was considered as an independent variable, while $d_{AT}$ was set to 20 cm. Four $\theta_{OBS}$ values, 35, 40, 45, and 50 degrees, were applied to the controller. The results of this experiment are shown in Fig. \ref{fig:kangle}. It illustrates the relation between the $\theta_{OBS}$ angle and the curve of reaction when the robot enters the avoidance zone. The bigger $\theta_{OBS}$ is, the larger is the human-robot security distance. 
\subsection{Avoidance threshold distance $d_{AT}$}
\begin{figure}[h!]
  \centering
  \includegraphics[width=0.44\textwidth]{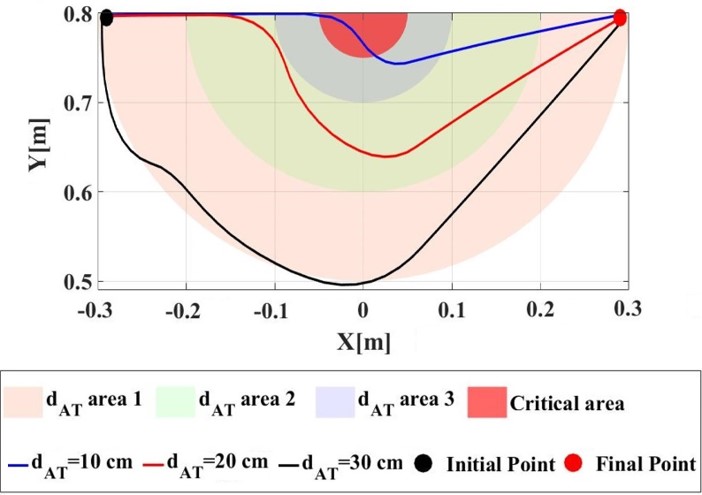}
  \qquad
  \caption{Experiment 2: the end-effector path curvature in relation to $d_{AT}$ values, plane (XY) view.}
  \label{fig:dmin}
\end{figure}

For this experiment, $d_{AT}$ was considered as an independent variable, while  $\theta_{OBS}$ was set to 45 degrees. Three $d_{AT}$ values, which are 10, 20, and 30 cm, were used to generate three different avoidance areas. The results of this experiment are shown in Fig. \ref{fig:dmin}. It explains how the robot's trajectory was affected by the avoidance controller when the TCP enters the avoidance region. It is important to notice that, a small $d_{AT}$ (10 cm) does not permit a proper evasion, on the other hand, a big $d_{AT}$ (30 cm) generates an unnecessary delay and displacement of the robot.

\section{Algorithm Evaluation Experiments}

In this section, the algorithm has been implemented in a real-time Human-Robot Interaction (HRI) application to evaluate the system performance. The user was located in front of the collaborative robot UR10 and was asked to wear the HTC tracker on the right hand, as shown in Fig. \ref{fig:SystemOverview}. The UR10 was continuously moving through three points, creating a virtual triangle. The input for the collision avoidance algorithm is the distance between the tracked human hand and the robot TCP. The experiment evaluates the algorithm's behavior when the robot and the operator have the same working area.
   
\subsection{Experimental Parameters}

In this experiment, the avoidance threshold distance, critical avoidance threshold distance, deactivation critical threshold distance, and  obstacle threshold angle are 0.2 m, 0.1 m, 0.2 m, and 45 degrees, respectively.

\subsection{Results}
\begin{figure}[h!]
  \centering
  \includegraphics[width=0.42\textwidth]{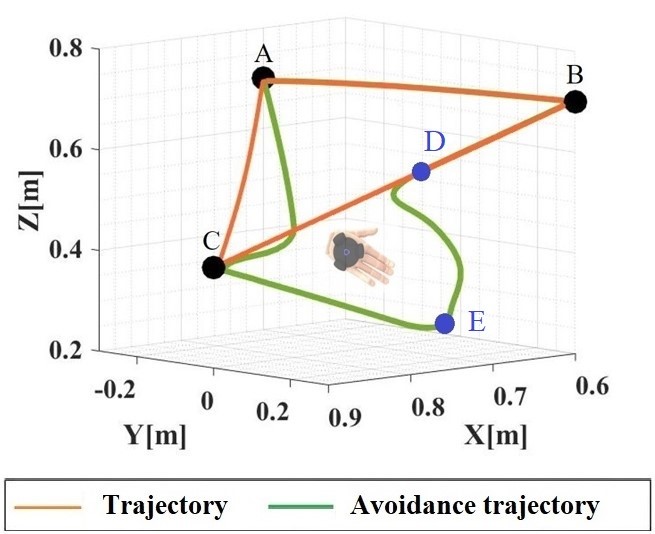}
  \qquad
  \caption{TCP trajectories. The orange trajectory represents the robot movement without obstacles. The green line illustrates the trajectory with random environmental obstacles. The A, B, and C points are the target positions. D-E is the section of the trajectory where the obstacle avoidance algorithm works. }
  \label{fig:Ex1}
\end{figure}

\begin{figure}[h!]
  \centering
  \includegraphics[width=0.48\textwidth]{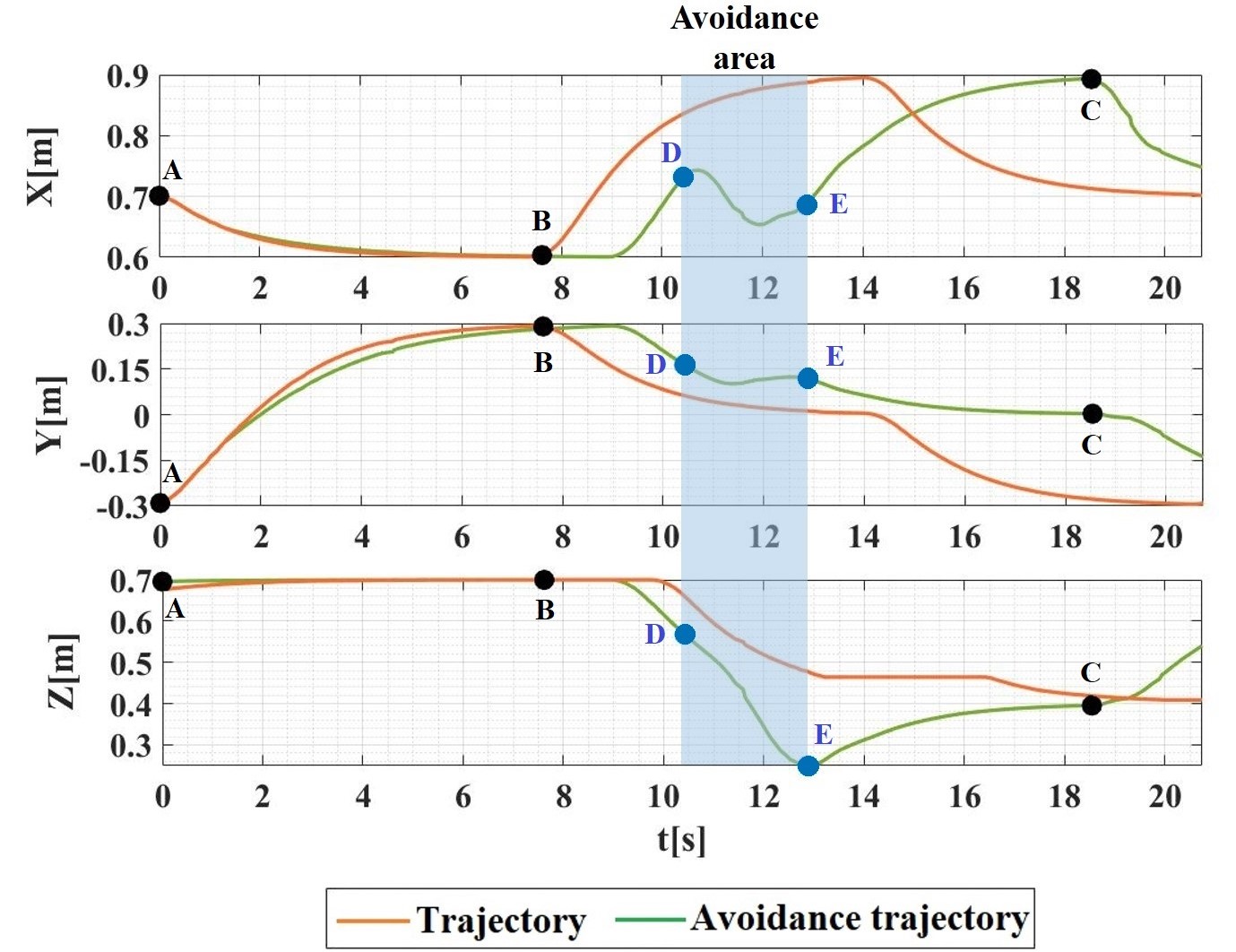}
  \qquad
  \caption{The diagram represents the trajectories with and without obstacle presence. The A, B, and C points are the target positions. D-E is the section of the trajectory where the obstacle avoidance algorithm works. }
  \label{fig:Ex2}
\end{figure}

The results of the first scenario are shown in Fig. \ref{fig:Ex1}. The orange line shows the robot TCP trajectory without collision avoidance following the path between the positions A, B, and C. The green line shows the modified trajectory  in the presence of the user's hand. The trajectories of the TCP illustrating the behavior of the proposed algorithm are presented in Fig. \ref{fig:Ex2}.

The duration of the task execution increases when the robot avoids the user's hand. When the user locates hand near to the robot working area, i.e. in the B-C and C-A trajectory sections, the algorithm avoided the hand and achieved the next goal point successfully. The blue highlighted area in Fig.  \ref{fig:Ex2} shows period when the collision avoidance algorithm is active (when $d_{RO} < d_{AT}$).

\section{Human-Robot Interaction Experiment in a Collaborative Task}

After calibrating the controller and evaluating the system on a preliminary experiment, a new proposed task evaluates the algorithm when the user and the robot perform a collaborative activity in a shared working environment. This experiment resembles common activities in a warehouse. The task objective was to put objects (cubes and cylinders) from a rack into boxes. The user was located in front of the collaborative robot UR10 and was asked to wear the AntiLatency 6 DOF camera on the right hand.

\begin{figure}[h!]
  \centering
  \includegraphics[width=0.48\textwidth]{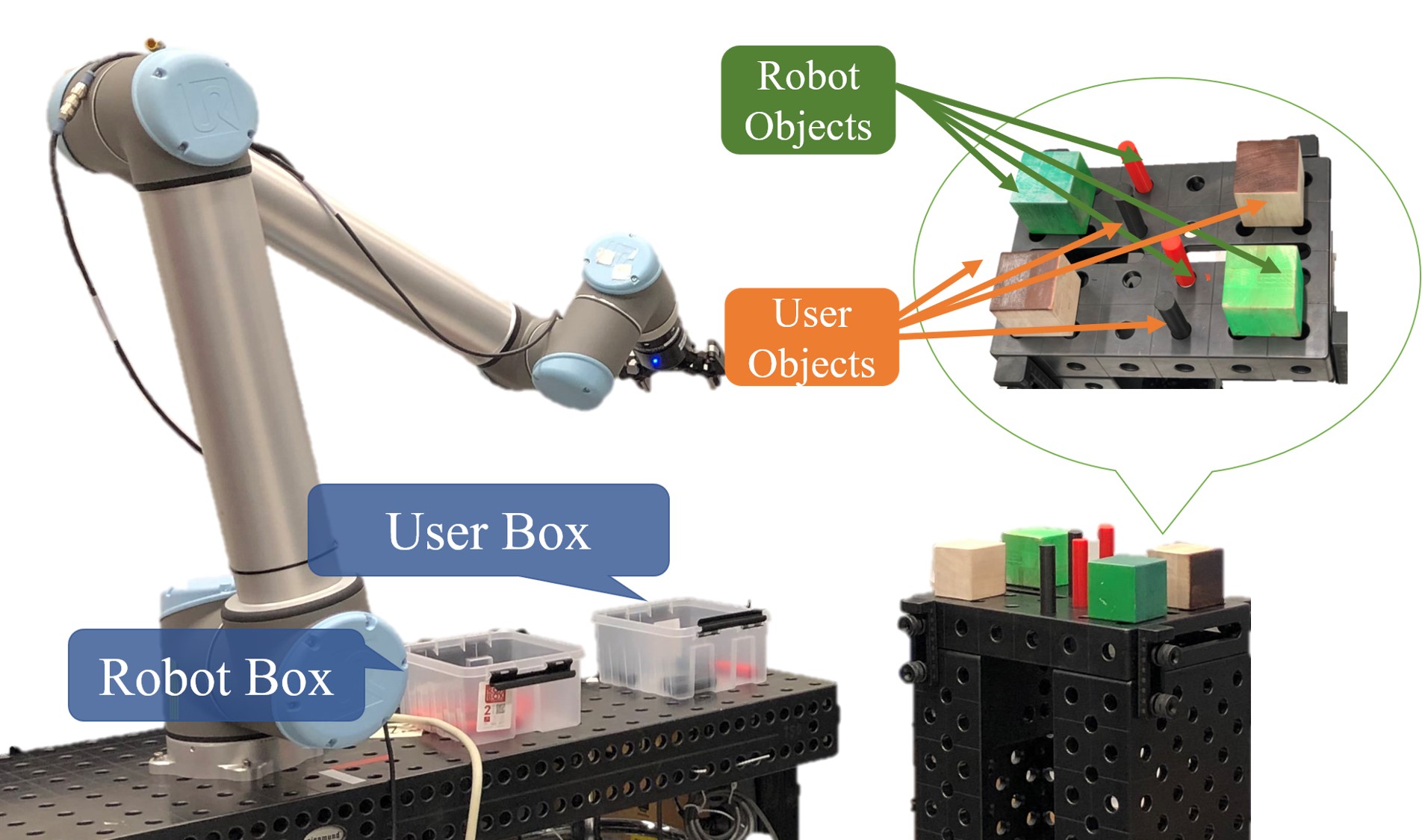}
  \qquad
  \caption{Experimental setup. Robot takes green cubes and red cylinders. The user takes brown cubes and black cylinders. Then they place cubes and cylinders in their respective boxes. }
  
  \label{fig:EnviromentExp3}
\end{figure}

The robot took two cubes and two cylinders from the rack and moved them to a box. The human did the same activity with the similar objects. The position of the objects in the environment is shown in Fig. \ref{fig:EnviromentExp3}. The goal of the experiment is to evaluate the proposed algorithm under conditions of a high probability of collision.

\subsection{Experimental Parameters}

In this experiment, the avoidance threshold distance, critical avoidance threshold distance, deactivation critical threshold distance, and  obstacle threshold angle are 0.2 m, 0.1 m, 0.2 m, and 45 degrees, respectively.

\subsection{Results}

\begin{figure}[h!]
  \centering
  \includegraphics[width=0.48\textwidth]{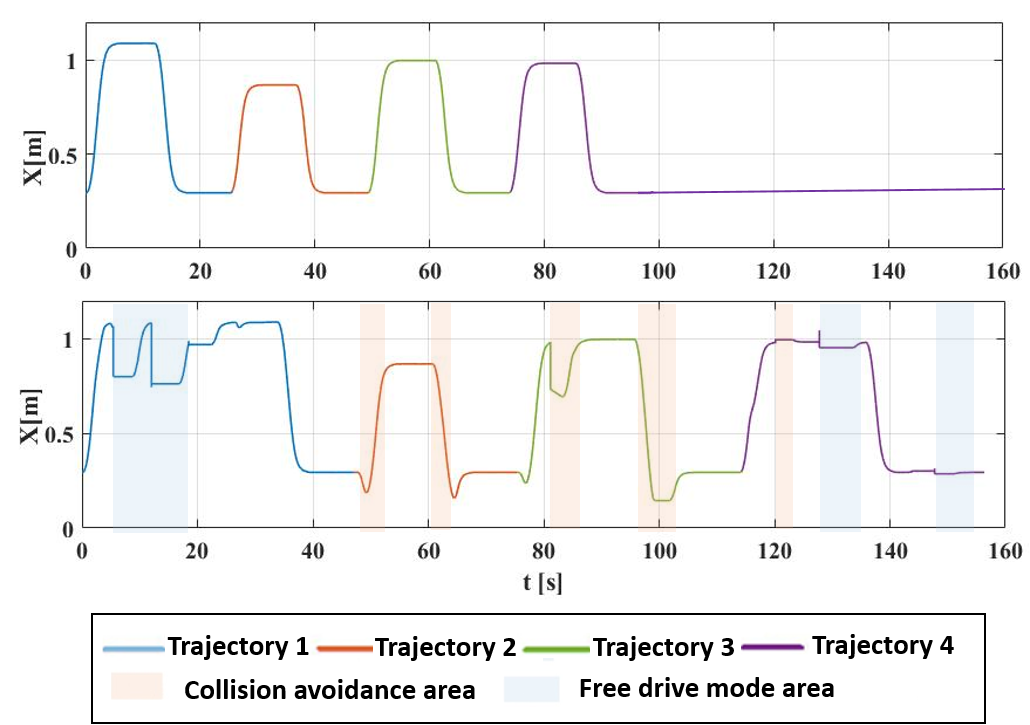}
  \qquad
  \caption{The diagram represents the trajectories without obstacle (top) and with obstacles (down) vs. time. The trajectories 1, 2, 3, and 4 represent TCP path to pick up each target (cube 1, cube 2, cylinder 1, and cylinder 2) and to transport them to the collecting box. The areas highlighted in orange and blue show the collision avoidance and free drive control mode cases, respectively. }
  \label{fig:3axis2}
\end{figure}

During the analysis of the results of the second experimental evaluation, four trajectories followed by the robot during the task are compared and inspected. In Fig. \ref{fig:3axis2}, the trajectories of TCP vs. time are presented. In the first part of the figure (top), the TCP trajectories without obstacles are shown, while the second part (down)  shows the TCP trajectories avoiding the user's hand. 

The activation of the collision avoidance mode is represented as orange highlighted areas (see Fig. \ref{fig:3axis2}). When the operator's hand enters the robot working area, the avoidance algorithm is activated, sidestepping the obstacle and, after that, immediately returning to the normal trajectory. As expected, the duration of the task increases when collision avoidance is activated due to a prolonged trajectory. Nevertheless, the robot and the user did not experience any collision. Thus, the proposed system allows a collision-free human-robot collaborative environment.

The activation of the free drive control mode is shown in Fig. \ref{fig:3axis2} (down) as blue highlighted area. This mode is activated when the distance to the hand is less than the critical avoidance area. In this mode, the user can modify the robot's position and orientation. It was observed that this mode allowed to take out the robot from the user's workspace by pulling-pushing the robot TCP, enhancing the interaction in a collaborative environment.

\section{Conclusions and Future Work}

In this paper, a novel approach to safe human-robot collaboration using wearable optical motion capturing systems was introduced, applying 6 DOF wearable optical mocap (by AntiLatecy), with the advantage that the working are can be easily arranged, and, if the camera is occluded, the hand position is available thanks to IMU odometry.  A novel collision avoidance method, based on APF for human-robot interaction (HRI) had been proposed.

The algorithm is based on the identification and discrimination of two types of obstacles. Besides, it was validated how calibration parameters modify the response. In the experiments, the parameters were $d_{AT}$ = 0.2 m and $\theta_{OBS}$ = 45 degrees. During the experiments, the robot governed by the proposed algorithm did not collide with the user. The introduced technology allows the user to work and interact with a robot in the same environment safely.

In future work, the velocity of the obstacle will be analyzed to improve the algorithm response time. Additional tacking points on the human limbs will be included in the system to enhance the safety.

The proposed technology will potentially allow humans to safely interact with multiple robots in cluttered environments, e.g., factories, kitchen, living room, and restaurants. In one scenario, visitors of the shop could wear cameras, IR markers, or QR codes. Wearable cameras or robots would track hand positions to avoid the collision and to perform physical HRI tasks, e.g., taking the plate from the client.

\section*{Acknowledgement}
The authors would like to thank AntiLatency for providing motion capture system equipment.

\addtolength{\textheight}{-12cm}   





\end{document}